\documentclass[10pt,twocolumn,letterpaper]{article}

\usepackage[pagenumbers]{cvpr}      
\definecolor{cvprblue}{rgb}{0.21,0.49,0.74}
\usepackage[pagebackref,breaklinks,colorlinks,allcolors=cvprblue]{hyperref}
\usepackage{url}
\usepackage{graphicx}
\usepackage{booktabs}
\usepackage{multirow} 
\usepackage{adjustbox}
\usepackage{tabularx}

\def\paperID{****} 
\def\confName{CVPR}
\def\confYear{2026}

\title{Aerial World Model for Long-horizon Visual Generation and \\Navigation in 3D Space}

\author{\textbf{Weichen Zhang}$^{*}$, \textbf{Peizhi Tang}$^{*}$, \textbf{Xin Zeng}$$, \textbf{Fanhang Man}$$, \textbf{Shiquan Yu}$^{}$, \textbf{Zichao Dai}$$, \textbf{Baining Zhao}$$, \\\textbf{Hongjin Chen}$$, \textbf{Yu Shang}$$, \textbf{Wei Wu}$$, \textbf{Chen Gao}$$, \textbf{Xinlei Chen}, \textbf{Xin Wang}, \textbf{Yong Li}, \textbf{Wenwu Zhu}\\
Tsinghua University \\ $^*$Equal Contribution\\
}

\begin{document}
\maketitle
\begin{abstract}
Unmanned aerial vehicles (UAVs) have emerged as powerful embodied agents. One of the core abilities is autonomous navigation in large-scale three-dimensional environments. Existing navigation policies, however, are typically optimized for low-level objectives such as obstacle avoidance and trajectory smoothness, lacking the ability to incorporate high-level semantics into planning. To bridge this gap, we propose ANWM, an aerial navigation world model that predicts future visual observations conditioned on past frames and actions, thereby enabling agents to rank candidate trajectories by their semantic plausibility and navigational utility. ANWM is trained on 4-DoF UAV trajectories and introduces a physics-inspired module: Future Frame Projection (FFP), which projects past frames into future viewpoints to provide coarse geometric priors. This module mitigates representational uncertainty in long-distance visual generation and captures the mapping between 3D trajectories and egocentric observations. Empirical results demonstrate that ANWM significantly outperforms existing world models in long-distance visual forecasting and improves UAV navigation success rates in large-scale environments.
\end{abstract}    
\section{Introduction}
\label{sec:intro}

\begin{figure*}[t]
  \centering
  \includegraphics[width=\linewidth]{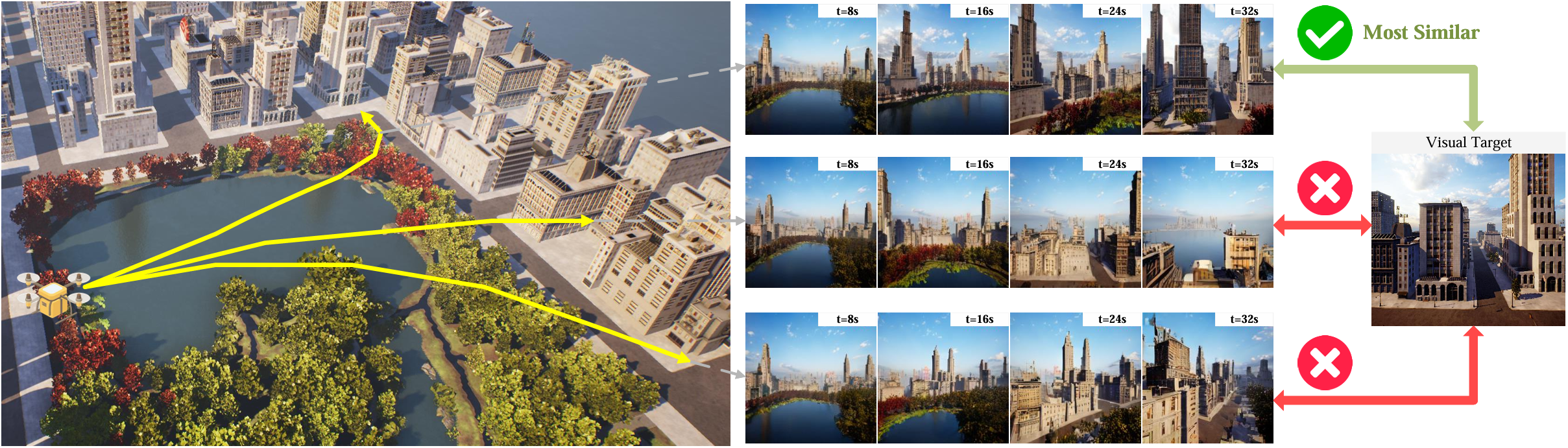}
  \vspace{-0.5cm}
  \caption{\textbf{Visual navigation in large-scale aerial space.} Given a visual target, the agent is required to plan a trajectory whose final observation aligns with the target. We leverage a world model that imagines visual observations along all possible trajectories. By computing the similarity between the imagined observations and the target, the optimal trajectory is determined. This imagination-based planning paradigm potentially reduces the navigation cost in large-scale open 3D environments.}
  \vspace{-0.4cm}
  \label{fig:story_overview}
\end{figure*}

Unmanned aerial vehicles (UAVs), as emerging intelligent agents, have demonstrated significant application value in the field of spatial intelligence~\cite{yang2025thinking}. 
One fundamental capability of UAV is visual navigation in 3D spaces, where the UAV needs to plan its path to search for a visual target efficiently. This capability is crucial for downstream tasks such as object search~\cite{gadre2023cows,zhang-etal-2025-citynavagent}, surveillance~\cite{semsch2009autonomous,ahmad2025future}, and logistics~\cite{chen2024ddl,chen2022deliversense}. 


Early works uses ``hard-coded'' navigation policies~\cite{shah2022gnm,sridhar2024nomad} to search the visual target in unseen environments. They primarily optimize for low-level objectives such as obstacle avoidance and path smoothness, but fails to integrate high-level semantics to facilitate path planning~\cite{sancaktar2025sensei,yokoyama2024vlfm}. 
Inspired by the human navigation ability to not only understand the current environment but also counterfactually imagine future scenarios without executing real actions~\cite{seeber2025human}, recent studies~\cite{mao2025yume,wang2023dreamwalker,wang2025dreamnav} have leveraged world models to imagine visual observations conditioned on future trajectories.
This enables the integration of semantic information about prospective scenes into path planning, supporting more efficient navigation.
However, existing methods~\cite{zhang2025matrix,yu2025gamefactory} remain limited to predicting short-horizon observations in 2D space.
For example, NWM~\cite{bar2025navigation} can only generate visual observations within a 3-meter range. While Genie 3~\cite{genie3} demonstrates strong long-horizon generative capabilities, its action space is constrained to the 2D plane.



In that case, constructing a world model for visual navigation in aerial spaces has two main challenges. 
1) \textbf{Complex action space.}
Compared to ground robots with only three degrees of freedom (DoF), UAVs have six DoF. Even without considering pitch and roll, the UAV action space remains four-dimensional.
Building a world model that can accurately map such a high-dimensional action space to corresponding visual observations is inherently difficult.
2) \textbf{Long-horizon visual generation.}
Unlike indoor navigation, aerial navigation typically involves long-horizon locomotion, where the visual target is usually beyond the current field of view and often over 100 meters away from the UAV.
Therefore, long horizon refers not only to the temporal dimension but also to the spatial extent.
Ensuring long-horizon spatial and temporal consistency in generated visual observations is particularly challenging.

To address the challenges above, we propose an Aerial Navigation World Model (ANWM) that predicts future visual observations conditioned on past observations and future trajectories.
To tackle the complex 3D action space, we introduce a 3D visual navigation benchmark that enables ANWM to learn the mapping from 3D actions to aerial observations.
To ensure long-horizon spatial–temporal consistency, we design a Future Frame Projection (FFP) module that projects past frames into future perspectives, enforcing visual consistency between the generated future observations and historical ones within their overlapping field of view.
Once trained, ANWM is used to predict visual observations along candidate trajectories generated by the path planning policy, enabling the agent to rank trajectories most likely to reach the target.

ANWM is conceptually related to recent diffusion-based world models for navigation and interactive tasks, such as NWM~\cite{bar2025navigation} and Matrix-Game~\cite{zhang2025matrix}. However, unlike these approaches, ANWM is specifically trained to generate first-person visual observations of aerial agents operating in large-scale 3D environments, which introduces unique challenges as discussed above. The main contributions of this paper are as follows:
\begin{itemize}[leftmargin=*,partopsep=0pt,topsep=0pt]
\setlength{\itemsep}{0pt}
\setlength{\parsep}{0pt}
\setlength{\parskip}{0pt}
    \item We introduce a large-scale dataset for training and testing the world model for aerial visual generation and navigation, containing 350k trajectory segments with corresponding visual observations. 
    \item The first action-conditioned world model in aerial space, capable of predicting long-horizon visual observations from 3D actions.
    \item Experimental results demonstrate that our proposed ANWM exhibits spatio-temporally consistent capability in long-horizon navigation, outperforming existing action-conditioned world models.
\end{itemize}

\section{Related Work}
\label{sec:related_work}

\textbf{Interactive World Models.}
Recent advances in world modeling have sought to endow agents with a unified representation of perception, action, and prediction within dynamic environments. Early generative approaches~\cite{peebles2023scalable,yang2024cogvideox,wan2025wan,esser2024scaling,kong2024hunyuanvideo} such as iVideoGPT~\cite{wu2024ivideogpt}, MineWorld~\cite{guo2025mineworld} demonstrated that large video transformers can implicitly capture physical dynamics and object interactions from raw visual sequences, laying the foundation for learning predictive simulators~\cite{chung2023luciddreamer,yu2024wonderjourney,yu2025wonderworld,liang2025wonderland} from pixels. 
Building upon this, GAIA-1~\cite{hu2023gaia} and GAIA-2~\cite{russell2025gaia} introduced large-scale multimodal world models capable of performing tool-augmented reasoning and web-based information synthesis, moving beyond static simulation toward \emph{interactive reasoning}. 
These models highlight a transition from passive next-frame prediction to active inference—where the agent continuously updates its internal world model based on feedback, search results, or user-provided context. 
Recent works~\cite{zhang2025matrix,yu2025gamefactory} such as YUME~\cite{mao2025yume}, and Genie 3~\cite{genie3} further extend this paradigm by incorporating generative imagination and high-fidelity visual synthesis, enabling real-time interaction and controllable environment simulation that bridge the gap between embodied intelligence and creative reasoning.

\noindent\textbf{World Models for Navigation.}
The core insight of leveraging world models for navigation lies in generating observations of unobserved scenes, enabling the agent to perform look-ahead planning for future trajectories.
PathDreamer~\cite{koh2021pathdreamer} first introduced this idea by generating panoramic observations for indoor waypoint prediction. Follow-up methods such as DreamWalker~\cite{wang2023dreamwalker}, DreamNav~\cite{wang2025dreamnav}, and UniWM~\cite{dong2025unified} further enhanced long-term planning through future-scene imagination.
To reduce the complexity of pixel-level generation, NavMorph~\cite{yao2025navmorph} predicts future latent world states instead of images, while HNR~\cite{wang2024lookahead} employs NeRF-based latent representations for efficient semantic encoding. Besides, TextDreamer~\cite{zhang2025cross}, use textual state representations to improve semantic abstraction.
Beyond next-step prediction, NWM~\cite{bar2025navigation} and DreamNav~\cite{wang2025dreamnav} generate global future observations conditioned on entire candidate trajectories, supporting global planning.
In parallel, panoramic world models like PanoGen~\cite{li2023panogen} and WCGEN~\cite{zhong2024world} synthesize text-conditioned indoor environments to mitigate data scarcity in VLN benchmarks.

Despite these advances, existing approaches remain largely confined to indoor, 2D settings. Extending these world-model-based imagination and navigation frameworks to outdoor, large-scale, and 3D open spaces remains a significant open challenge.

\begin{figure*}[t!]
  \centering
  \includegraphics[width=\linewidth]{images/framework_v3.pdf}
  \vspace{-0.8cm}
  \caption{\textbf{The Framework Overview.} a) We collect the datasets from AVLN simulators and generate trajectory clips by action enrichment and random partition. 
  b) For single-frame generation, ANWM produces future visual observations conditioned on the noisy latent, the past-frame latent, the projected future-frame latent, and the embedding of the upcoming action. We employ the Future Frame Projection module to warp the past frame into the future viewpoint, providing a strong scene prior for generation. c) For long-horizon generation, ANWM operates in an autoregressive manner to generate sequential visual observations along the trajectory. Each newly generated frame is appended to the past-frame queue which is then used as input for the next observation generation.}
  \vspace{-0.4cm}
  \label{fig:frame_overview}
\end{figure*}

\section{ANWM: Aerial Navigation World Model}
\label{sec:method}
\subsection{Formulation}

In this section, we describe the formulation of visual navigation settings and ANWM. 

In the visual navigation task, an agent is required to search for a target specified by an image. The objective is to navigate to a position where the agent’s visual observation most closely resembles the target image. Note that the target may or may not be visible from the agent’s current viewpoint, which distinguishes this problem from conventional path planning in robotics. In this case, the visual navigation problem can be formally formulated as follows.

Given the agent's current egocentric observation $\mathbf{v}_t{\in} \mathbb{R}^{H\times W\times 3}$ and its action space $\mathbb{A}$, the agent plans its future actions $\mathbf{D}{=}({ \mathbf{a}_{t+1}, ..., \mathbf{a}_n})$ to reach a location with a final observation $\mathbf{v}_n$ closely resembles the target image $\mathbf{v}^*$. $\mathbf{a}_k{\in}\mathbb{A}$ is the basic action command of the agent given by relative translation $(\Delta x, \Delta y, \Delta z){\in}\mathbb{R}^3$ and yaw change$\Delta \varphi{\in} \mathbb{R}$. The objective can be formulated as:
\begin{equation}
\label{eq:eq1}
\begin{split}
    \mathbf{D}^*= (\mathbf{a}^*_{t+1},\dots,\mathbf{a}^*_{n}) &= \arg\max_{\mathbf{D}\in\mathbb{D}}\ \mathcal{S}(\mathbf{v}_n, \mathbf{v}^*)\\
    \text{ s.t. } \mathbf{v}_n&=\mathcal{F}(\mathbf{D},\mathbf{v}_t)
\end{split}
\end{equation}
where $\mathcal{S}:(\mathbf{v}_{i}, \mathbf{v}_{j}) {\mapsto} \mathbb{R}$ is the similarity scores between two latent states, $\mathbb{D}$ is the set of agent's possible trajectories and $\mathcal{F}$ is the agent's kinematic model.
Since exhaustively exploring all possible trajectories in open environments would incur prohibitive navigation time and costs, ANWM aims to exploit the counterfactual reasoning capability of generative world models. It enables the agent to \emph{imagine} future observations without executing real actions as depicted in Figure~\ref{fig:story_overview}. 
Therefore, the objective of ANWM is to learn a world model $W$ that accurately simulates the distribution of observations conditioned on historical observations and actions:
\begin{equation}
\label{eq:eq3}
    \mathbf{v}_{k+1} \sim W_{\theta}(\mathbf{v}_{k+1}\mid \mathbf{v}_k, a_{k+1})
\end{equation}
where $\theta$ denotes the model parameters. 
We also assume a $m$-order Markov property in the agent's visual observations, such that the $\mathbf{v}_{k+1}$ depends only on the most recent $m$ observations $\mathbf{v}_{k-m:k}$. Accordingly, Equation \ref{eq:eq3} becomes:
\begin{equation}
\label{eq:eq4}
    \mathbf{v}_{k+1}\sim  W_{\theta}(\mathbf{v}_{k+1}\mid \mathbf{v}_{k-m:k}, a_{k+1})
\end{equation}



Thus, ANWM can generate all observations along the trajectory in an autoregressive manner. By comparing the final generated observation with the target observation, ANWM selects the trajectory with the highest similarity score as the navigation path to be executed.

\subsection{Dataset for Aerial Navigation World Model}
\label{subsec:data_gen}
We first present the egocentric aerial agent video dataset along with will-aligned trajectories for both training and testing. As depicted in Figure~\ref{fig:frame_overview}(a), we first collect UAV trajectories from the aerial vision-and-language navigation (AVLN) benchmarks, including AeralVLN~\cite{liu2023aerialvln}, OpenFly~\cite{gao2025openfly}, and OpenUAV~\cite{wang2024towards}. These benchmarks contain more than 20k diverse aerial trajectories, spanning over 40 simulated urban scenes built on Unreal Engine~\cite{unrealengine}. Each trajectory is represented by a sequence of 3D waypoints and paired with a language instruction. 
We replay the UAV’s flight along each trajectory in Unreal Engine to obtain the temporal RGB-D observations. 
Since the original trajectories are biased toward forward actions, we design an action enrichment strategy to mitigate this issue. During trajectory replay, we record not only the front-view images but also images from the left, right, and rear views. With this strategy, a forward motion in the front view can also be interpreted as a lateral movement in the side views or a backward motion in the rear view.



Finally, we collect 350k trajectory segments for training and 2.2k for testing, consisting of 1.1k 2D and 1.1k 3D segments.
Each segment includes 48 actions chosen from forward/backward, left/right, up/down, or left/right rotation, and only the post-action frame is recorded, yielding 48 frames per segment.
The UAV velocities are set to 5 m/s, 2 m/s, and 15°/s for horizontal, vertical, and rotational movement, resulting in an average path length of 80.7 meters.

\subsection{Navigation Framework Overview}
\textbf{World Model for Future Frame Generation}. As illustrated in Figure~\ref{fig:frame_overview}, the world model conditions on the past $m$ frames together with the next action to denoise the future frame into a physically plausible image. 
A pretrained VAE encoder~\cite{blattmann2023stable} is used first to compress the raw frames into latent representations with an $8\times8$ downsampling factor. The condition latents and the noisy future latent are then jointly processed through $N$ Conditional Diffusion Transformer (CDiT) blocks. Finally, the denoised latent is decoded back into the pixel space using the VAE decoder~\cite{blattmann2023stable}.
To incorporate the next action $a \in \mathbb{R}^4$ as a condition signal, the action encoder projects $a$ into an action embedding $\nu_a \in \mathbb{R}^d$ using sine–cosine features. $\nu_a$ is subsequently passed through an adaptive layer normalization module to produce the scale and shift coefficients that modulate the CDiT blocks.
To enable the model to generate more realistic and longer-horizon future observations, we introduce the future frame projection (FFP) module that explicitly encodes cross-view consistency into the generation process. The FFP module projects the most recent historical frame into the viewpoint of the next frame to serve as an auxiliary frame. Similar to the past frames, this auxiliary frame is encoded by the VAE encoder into a condition latent, providing the CDiT blocks with prior information about the future observation. Detailed architecture designs are described in Section~\ref{sec:method}.

\noindent\textbf{Autoregressive Inference and Path Planning}. After training, we leverage the model to assist path planning for aerial visual navigation as depicted in Figure~\ref{fig:frame_overview}c). We first leverage a heuristic path planner to produce $l$ candidate trajectories by sampling actions from a predefined action set. Then, Gaussian noise is applied as perturbations to the waypoints along each trajectory to further enhance trajectory diversity. 
ANWM is leveraged to generate the visual observation at the endpoint of each trajectory to rank these trajectories.
During the generation phase, each waypoint pose $(x, y, z, \varphi)$ is quantized into the relative transformation with respect to the previous waypoint, expressed as $(\Delta x, \Delta y, \Delta z, \Delta \varphi)$, to align with the input format of ANWM. It predicts the next frame in an autoregressive manner, where each newly generated frame is appended to the history buffer, and the most recent $m$ frames are encoded as the state context for subsequent frame generation. Once the final frame of a trajectory is generated, the model evaluates the perceptual similarity $\mathcal{S}(v_n, v^*)$ between the predicted last frame $v_n$ and the target frame $v^*$ using LPIPS. Among all candidate trajectories, the one with the lowest similarity error is then selected as the final navigation path. By default, the number of candidate trajectories $l$ is set to 5.


\begin{figure*}[t!]
  \centering
  \includegraphics[width=0.95\linewidth]{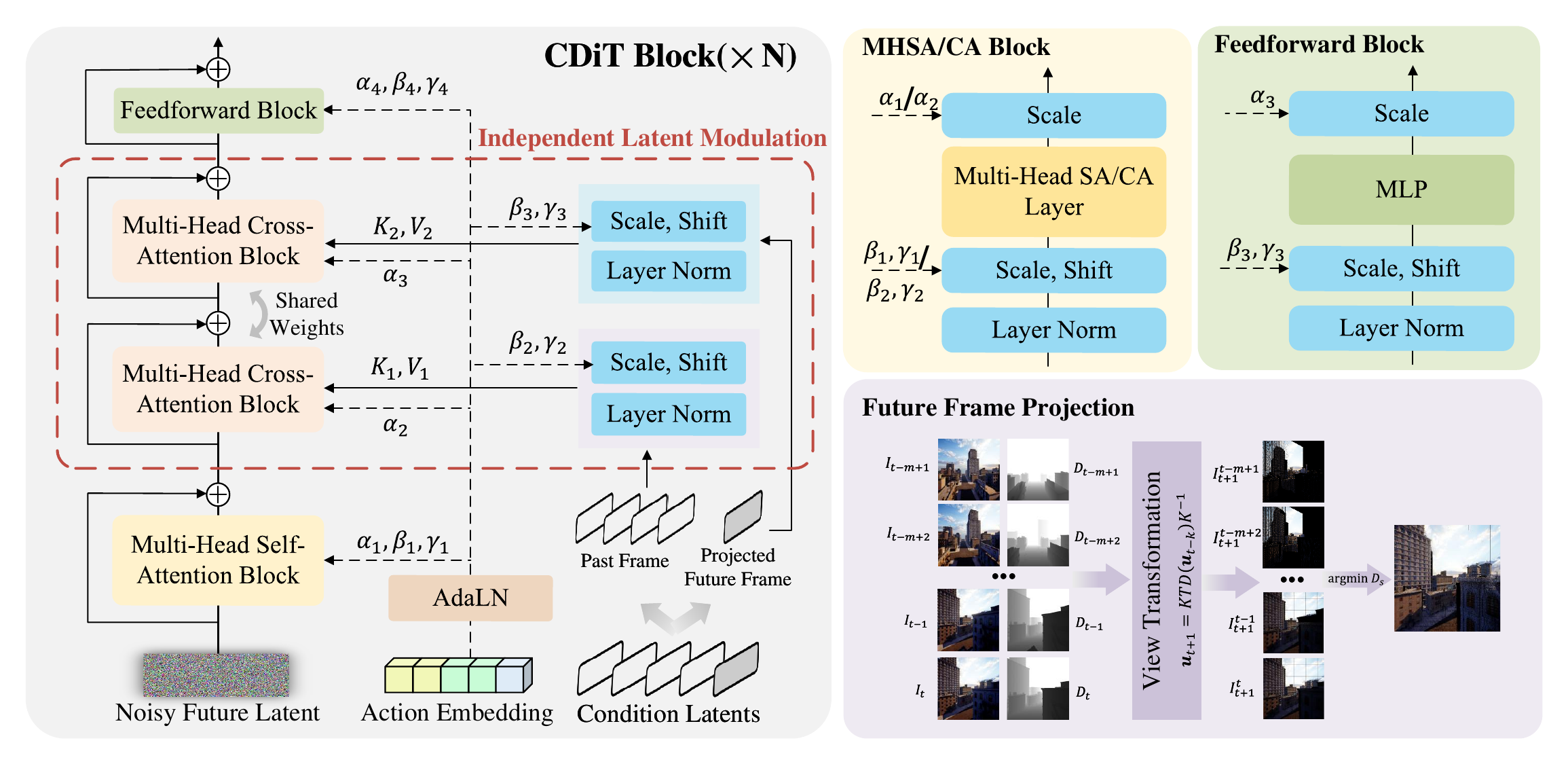}
  \vspace{-0.5cm}
  \caption{\textbf{Model Architecture.} ANWM adopts CDiT~\cite{bar2025navigation} as the backbone but uses the past frame and the projected future frame as distinct conditional signals to control the generation process. Specifically, ANWM first splits the condition latents into the past-frame latent and the projected future-frame latent, and applies separate scale and shift parameters to modulate the strength of the conditioning signal. The modulated latents are then fed into two shared-weight Multi-Head Cross-Attention branches.}
  \vspace{-0.4cm}
  \label{fig:arch_overview}
\end{figure*}

\subsection{Model Architecture}
\label{subsec:model_arch}

\noindent\textbf{Future Frame Projection (FFP).}
Rather than directly feeding the past frame latents into the world model, we design the FFP module that generates a coarse future frame prior, which is then concatenated with the past frame latents as the conditional input. This module leverages the view transformation method in 3D vision that projects $m$ past frames $I_{t{-}m+1:t}$ into the UAV's future viewpoint at time $t+1$ to obtain an estimated target frame $\tilde{I}_{t+1}$. 
Given a source frame $I_{t-k}$ with depth $D_{t-k}$, camera intrinsics $K$, and relative pose $T_{t-k \rightarrow t{+}1}$, each pixel $\mathbf{u}=[x,y,1]^\top$ is first back-projected into 3D space via: 
$ \mathbf{p}(x,y) = D_{t-k}(x,y)\,K^{-1}\mathbf{u}$.
Then, the 3D point $\mathbf{p}$ is transformed into the target image plane via
$\tilde{\mathbf{u}} = KT_{t-k\rightarrow t{+}1}\,\mathbf{p}$, where $\tilde{\mathbf{u}}=[\tilde{x}\tilde{z}, \tilde{y}\tilde{z}, \tilde{z}]$. 
Using this view transformation function, we project the pixels in $I_{t-k}$ into the target frame: $I^{t-k}_{t{+}1}(\tilde{u}, \tilde{v})=I_{t-k}(u, v)$. 
Each projected target frame $I^{t-k}_{t+1}$ contains only a subset of pixels in the target frame and exhibits a substantial missing pixels.
To obtain a more complete estimation of the future frame, we fuse all projected frames $I^{t-m+1:t}_{t+1}$ into a single final target frame to compensate for these missing pixels.
Specifically, among all projected frames, we select the pixel value corresponding to the minimum depth across frames as the final pixel value of the target frame, which is given by:
\begin{equation}
\begin{split}
    \tilde{I}_{t+1}(x, y) &= I^{t-s^*(x,y)}_{t+1}(x,y),\\ s^*(x,y) &= \arg\min_{s\in [0:m-1]}D_s(x,y).
\end{split}
\end{equation}
In that case, we obtain a coarse estimation of the target frame by leveraging the visual cues from all historical observations, which provide an essential prior for future frame generation. 

\noindent\textbf{Independent Latent Modulation.} After obtaining the estimated future frame $\tilde{I}_{t+1}$, it is encoded together with the past frames $I_{t-m:t}$ by the VAE to form the conditional latents $x_{\text{cond}}= \text{VAE}([I_{t-m:t};\tilde{I}_{t+1}]) = ([x_{t-m:t};\tilde{x}_{t+1}]) \in \mathbb{R}^{B \times (m+1) \times C \times H \times W}$ which serves as the control signal for future frame generation. 
We use the Conditional Diffusion Transformer (CDiT)~\cite{bar2025navigation} as our aerial world model backbone. 
As depicted in Figure~\ref{fig:arch_overview}, given an input of noisy future latent $x_{t+1}'\in \mathbb{R}^{B\times C \times H \times W}$, condition latents $x_\text{cond}$ and an action embedding $\nu_a$, CDiT model predicts the denoised future latent $x_{t+1}$ by applying $N$ CDiT blocks over the input latents, where $B$ and $C$ are batch size and channels. In each CDiT block, $\nu_a\in \mathbb{R}^{d}$ is used to generating scale $\boldsymbol{\alpha}\in \mathbb{R}^{4\times d_e}$, $ \boldsymbol{\beta} \in \mathbb{R}^{5 \times d_e}$ and shift $\boldsymbol{\gamma} \in \mathbb{R}^{5 \times d_e}$ coefficients by AdaLN~\cite{xu2019understanding} block:
\begin{equation}
   \boldsymbol{\alpha}, \boldsymbol{\beta}, \boldsymbol{\gamma} = \text{AdaLN}(\text{SiLU}(\nu_a)),
\end{equation}
where $d_e$ is the coefficient dimension. Although $x_{t-m:t}$ and $\tilde{x}_{t+1}$ both represent the agent’s observations at different time steps, they exhibit distinct feature distributions. The past frames are real observations and always semantically meaningful, while the projected future frame is a synthesized image with projection errors and can even become meaningless if no overlapping field of view between the two perspectives.
Therefore, we propose the Independent Latent Modulation (ILM) method to modulate the distributions of $x_{t-m:t}$ and $\tilde{x}_{t+1}$ separately. Specifically, these latents are passed into two separate modulation layers:
\begin{equation}
\begin{split}
    z_{t-m:t} &= (1+\beta_{2})\text{LN}(x_{t-m:t}) + \gamma_2, \\
    \tilde{z}_{t+1} &= (1+\beta_3)\text{LN}(\tilde{x}_{t+1})+\gamma_3,  \\
\end{split}
\end{equation}
The modulated condition latents are fed into two shared-weight MHCA blocks sequentially, which is given by:
\begin{equation}
\begin{split}
    z'_{t+1} &{=} x'_{t+1}{+}\alpha_1\text{MHSA}((1+\beta_1)\text{LN}(x'_{t+1})+\gamma_1), \\
    z''_{t+1} &{=} z'_{t+1}{+}\alpha_2\text{MHCA}(Q_1{=}z'_{t+1},K_1{=}V_1{=}z_{t-m:t}), \\
    z'''_{t+1} &{=} z''_{t+1}{+}\alpha_3\text{MHCA}(Q_2{=}z''_{t+1},K_2{=}V_2{=}\tilde{z}_{t+1}). \\
\end{split}
\end{equation}
Finally, the intermediate latent is processed by a feedforward block to produce the denoised latent $z_{t+1}$:
\begin{equation}
\begin{split}
    z_{t+1} {=} z'''_{t+1}+ \alpha_4\text{MLP}((1+\beta_4)\text{LN}(z'''_{t+1})+\gamma_4).  \\
\end{split}
\end{equation}

\begin{table*}[t]
\centering
\small
\caption{\textbf{Generative results} of 2D and 3D trajectories.}
\vspace{-0.2cm}
\label{tab:gen_2d3d}
\resizebox{0.9\linewidth}{!}{
\begin{tabular}{@{}llcccccc@{}}
\toprule
\multicolumn{2}{c}{} & \multicolumn{3}{c}{\textbf{2D}} & \multicolumn{3}{c}{\textbf{3D}} \\
\cmidrule(l){3-5} \cmidrule(l){6-8}
Timeshift & {Method} & \quad LPIPS $\downarrow$ & \quad DreamSim $\downarrow$ & \quad FID $\downarrow$ & \quad LPIPS $\downarrow$ & \quad DreamSim $\downarrow$ & \quad FID $\downarrow$ \\
\midrule
\multirow{4}{*}{\textbf{4s}} 
  & Matrix-Game  & 0.589 & 0.222 & 40.2 & - & - & - \\
  & YUME         & 0.571 & 0.196 & 73.0 & - & - & - \\
  & NWM          & 0.377 & 0.259 & 38.5 & 0.376 & 0.247 & 39.9 \\
  & ANWM (ours)  & \textbf{0.184} & \textbf{0.125} & \textbf{19.2} & \textbf{0.192} & \textbf{0.143} & \textbf{22.6} \\
\midrule
\multirow{4}{*}{\textbf{8s}} 
  & Matrix-Game  & 0.642 & 0.307 & 43.6 & - & - & - \\
  & YUME         & 0.694 & 0.375 & 143.1 & - & - & - \\
  & NWM          & 0.422 & 0.291 & 43.5 & 0.428 & 0.280 & 38.9 \\
  & ANWM (ours)  & \textbf{0.226} & \textbf{0.148} & \textbf{20.7} & \textbf{0.236} & \textbf{0.170} & \textbf{25.1} \\
\midrule
\multirow{4}{*}{\textbf{16s}} 
  & Matrix-Game  & 0.714 & 0.477 & 71.1 & - & - & - \\
  & YUME         & 0.853 & 0.701 & 232.3 & - & - & - \\
  & NWM          & 0.470 & 0.336 & 46.3 & 0.482 & 0.321 & 42.1 \\
  & ANWM (ours)  & \textbf{0.313} & \textbf{0.202} & \textbf{24.5} & \textbf{0.301} & \textbf{0.210} & \textbf{29.4} \\
\midrule
\multirow{4}{*}{\textbf{32s}} 
  & Matrix-Game  & 0.790 & 0.646 & 135.9 & - & - & - \\
  & YUME         & 0.902 & 0.787 & 269.7 & - & - & - \\
  & NWM          & 0.524 & 0.400 & 61.0 & 0.535 & 0.377 & 47.6 \\
  & ANWM (ours)  & \textbf{0.433} & \textbf{0.294} & \textbf{32.5} & \textbf{0.389} & \textbf{0.271} & \textbf{36.1} \\
\bottomrule
\end{tabular}
}
\end{table*}

\begin{figure*}[h]
  \centering
  \includegraphics[width=0.92\linewidth]{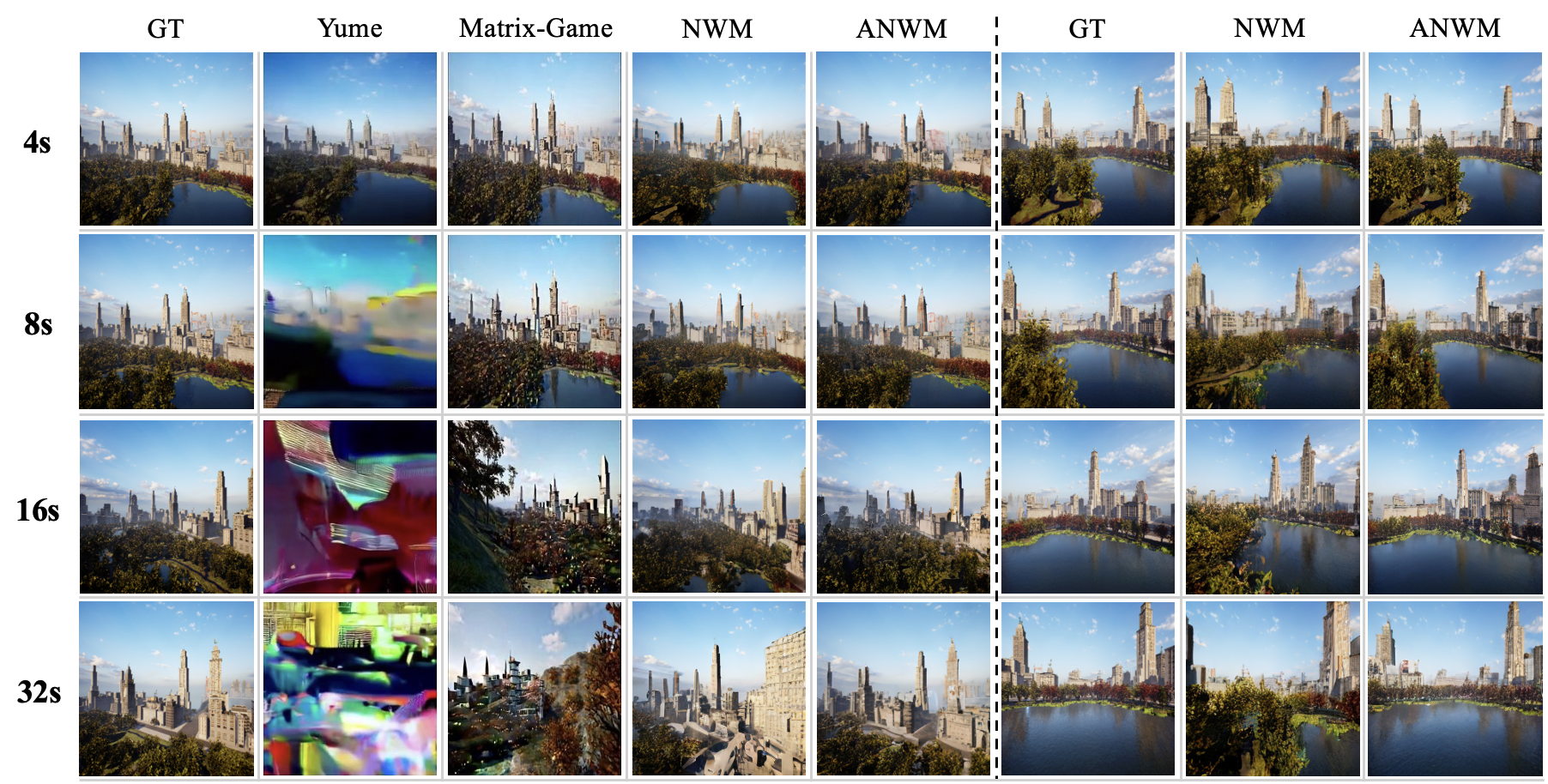}
  \vspace{-0.3cm}
  \caption{\textbf{The qualitative results of generative visual observation along the path.} Left: 2D trajectory. Right: 3D trajectory.}
  \vspace{-0.5cm}
  \label{fig:qual_res}
\end{figure*}

\section{Experiments}



\begin{table*}[t]
\centering
\small
\caption{2D and 3D navigation results.}
\vspace{-0.2cm}
\resizebox{\linewidth}{!}{
\begin{tabular}{@{}lcccccccc@{}}
\toprule
\multicolumn{1}{c}{} & \multicolumn{4}{c}{\textbf{2D}} & \multicolumn{4}{c}{\textbf{3D}} \\
\cmidrule(l){2-5} \cmidrule(l){6-9}
{Method} &  \qquad\  ATE $\downarrow$  &  \qquad\  RPE $\downarrow$ & \qquad\  SR $\uparrow$ &  \qquad\  NE $\downarrow$  & \qquad\  ATE $\downarrow$  &  \qquad\  RPE $\downarrow$  & \qquad\  SR $\uparrow$  & \qquad\  NE $\downarrow$ \\
\midrule
    YUME        & \quad\  15.92 & \quad\  1.80 & \quad\  0.0 & \quad\  24.32 & \quad\  - & \quad\  - & \quad\  - & \quad\   -\\
   Matrix-Game  & \quad\ 14.75 & \quad\ 1.53 & \quad\ 16.0 & \quad\ 20.98 &\quad\  - &\quad\  - &\quad\  - &\quad\  -\\
   NWM          & \quad\ 7.72 & \quad\ 0.89 & \quad\ 63.0 & \quad\ 12.71 & \quad\ 8.52 & \quad\ \textbf{1.03} & \quad\ 58.0 & \quad\ 14.51 \\
   ANWM (ours)  & \quad\ \textbf{6.30} & \quad\ \textbf{0.78} & \quad\ \textbf{73.0} &\quad\  \textbf{10.30} & \quad\ \textbf{8.13} & \quad\ 1.06 & \quad\ \textbf{60.0} & \textbf{14.12} \\
\bottomrule
\end{tabular}
}
\label{tab:main_nav}
\end{table*}

\begin{figure*}[h]
  \centering
  \includegraphics[width=\linewidth]{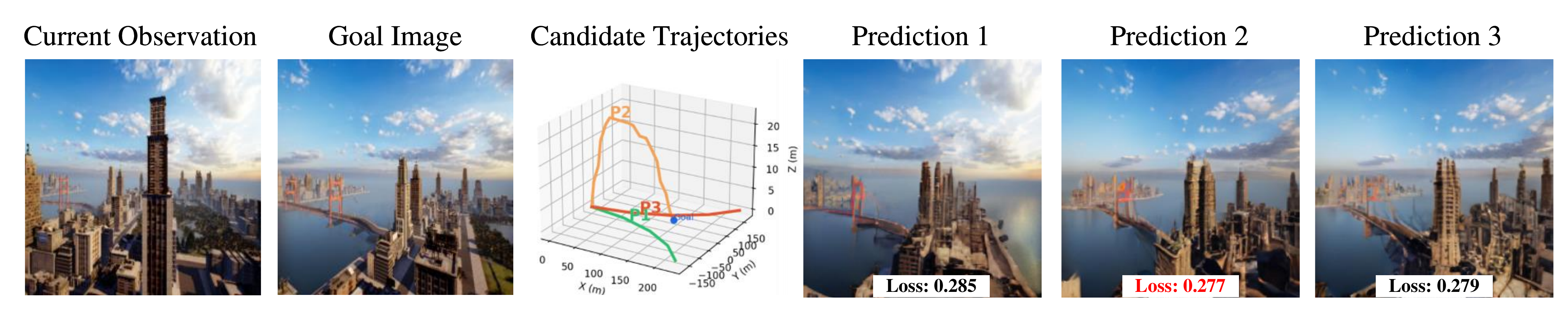}
  \vspace{-0.7cm}
  \caption{\textbf{The qualitative results of visual navigation.} ANWM ranks each trajectory's final prediction by measuring the LPIPS similarity with the goal Image. The trajectory with the lowest LPIPS is selected for execution. We only visualize the top-3 trajectories.}
  \vspace{-0.5cm}
  \label{fig:qual_nav}
\end{figure*}

\subsection{Experiment Setup}
\paragraph{Dataset and Benchmark}
We evaluate the performance using the 1.1k 2D trajectory segments and 1.1k 3D trajectory segments introduced in Section~\ref{sec:method}.
For both the 2D and 3D setups, 1,000 segments are used to test the model’s generative capability, and the remaining 100 segments are used to evaluate its navigation performance, which is consistent with the experimental setup of NWM~\cite{bar2025navigation}. For each 48-frame segment, we set the first 16 frames as historical observations, and the model is required to predict the next 32 frames. 

\noindent\textbf{Baselines}  We compare our method against three representative world models that generate future observations conditioned on action inputs: NWM~\cite{bar2025navigation}, Matrix-Game~\cite{zhang2025matrix}, and YUME~\cite{mao2025yume}.
Since the original architectures of these baselines only support 2D action inputs, we first compare their performance in visual generation and navigation on 2D trajectories.
For evaluation on 3D trajectories, we extend the action interface of NWM to accommodate 3D motion and retrain it on our dataset. Retraining Matrix-Game or YUME is not feasible because their source codes are not publicly available.
In addition, we compute the average motion velocity of the baseline agents and generate videos of varying durations to ensure that the distance traveled within the same time interval is consistent with that of ANWM.

\noindent\textbf{Metrics}
For the generation task, we employ FID~\cite{heusel2017gans}, DreamSim~\cite{fu2023dreamsim}, and LPIPS~\cite{zhang2018unreasonable} to evaluate the semantic fidelity of the generated results,
and use MSE, SSIM, and PSNR to assess their pixel-level accuracy. For the navigation task, we use Absolute Translation Error (ATE), Relative Pose Error (RPE)~\cite{sturm2012evaluating}, Success Rate (SR)~\cite{liu2023aerialvln}, and Navigation Error (NE)~\cite{gao2025openfly} to evaluate navigation accuracy. 

\noindent\textbf{Implementation Details.}
The input and output frames are resized to a resolution of $224 \times 224$. ANWM is implemented with 8 CDiT blocks and trained for 300k steps on four NVIDIA A800 40GB GPUs. We use the AdamW optimizer with a learning rate of $8e-5$. By default, the number of conditional past frames, also referred to as the context size $m$ is 4.
During inference, ANWM autoregressively generates the visual observation at each waypoint along the trajectory.

\begin{figure*}[t]
  \centering
  \begin{minipage}[t]{0.49\textwidth}  
    \vspace{0pt}
    \centering
    \includegraphics[width=\textwidth]{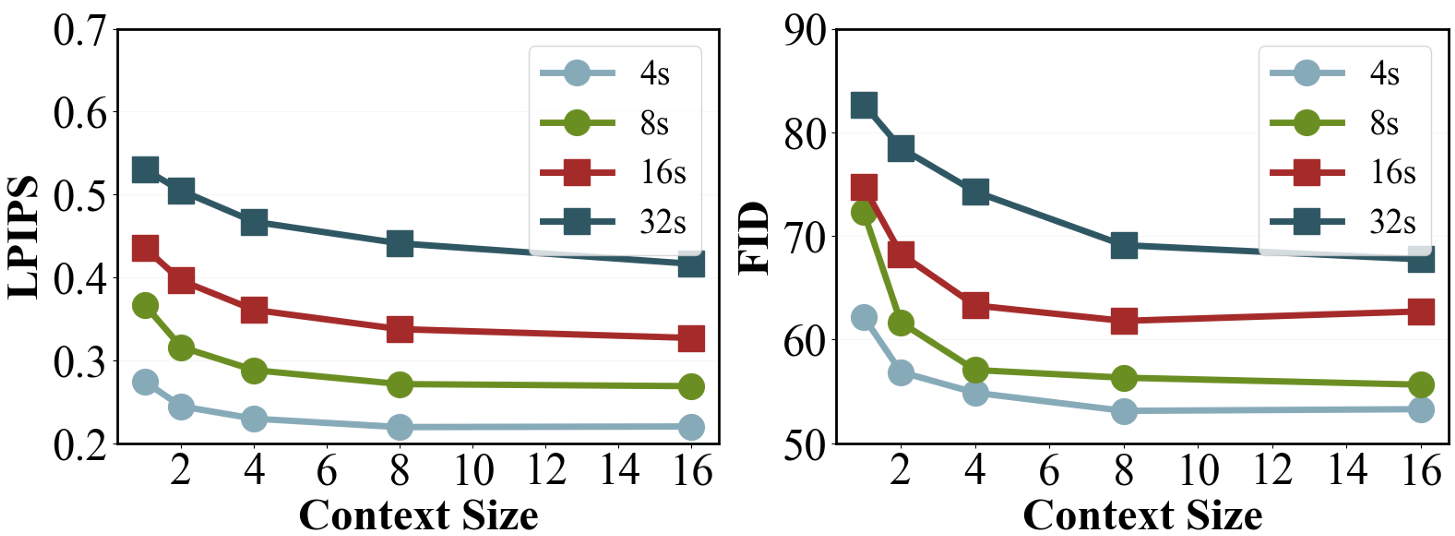}
    \vspace{-0.7cm}
    \caption{\textbf{Ablation of context size for future frame projection.}}
    \label{fig:ablation1}
  \end{minipage}
  \hfill
  \begin{minipage}[t]{0.24\textwidth}  
    \vspace{0pt}
    \centering
    \includegraphics[width=\textwidth]{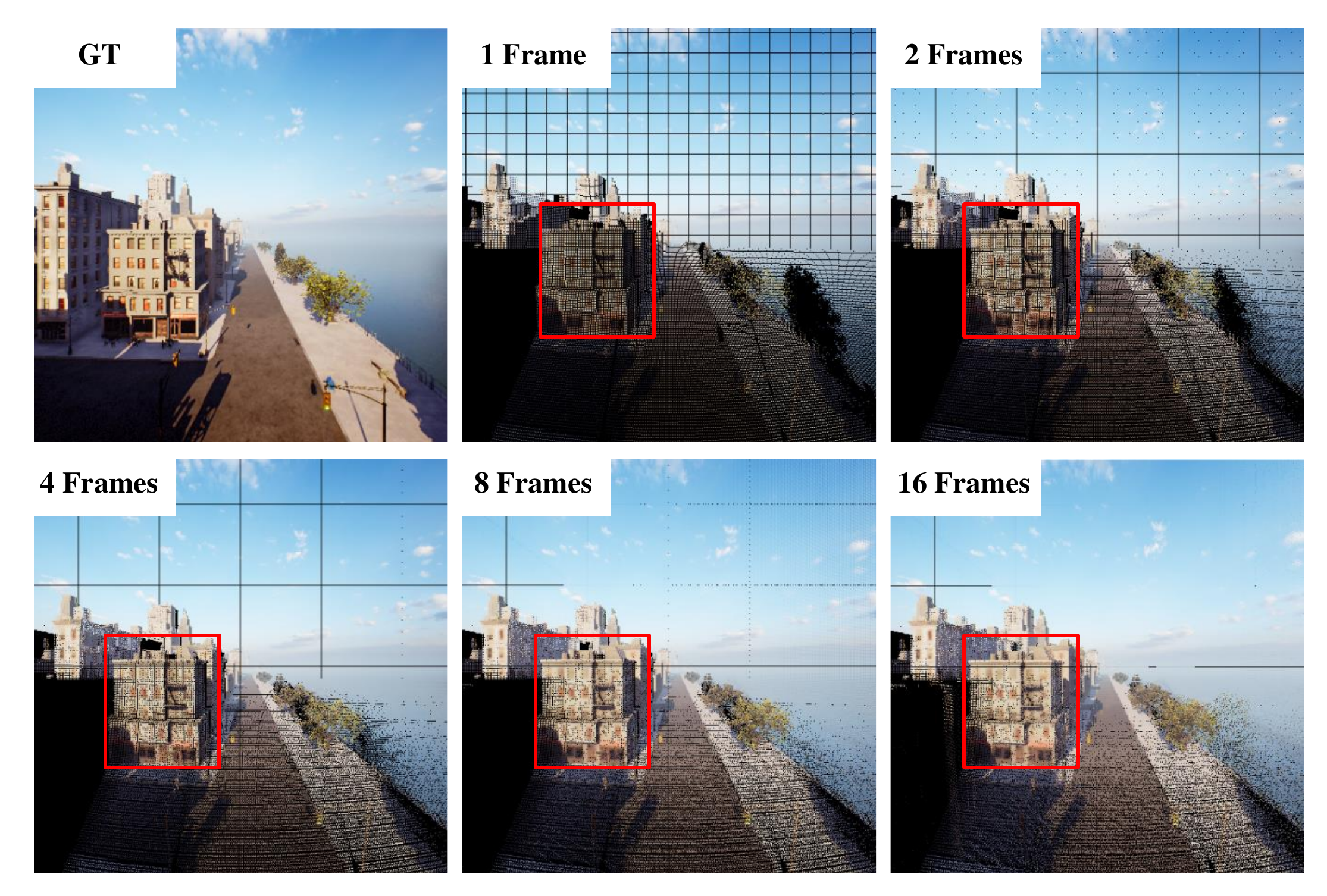}
    \vspace{-0.7cm}
    \caption{\textbf{Qualitative results of FFP via different context sizes.}}
    \label{fig:ablation1_qual}
  \end{minipage}
  \hfill
  \begin{minipage}[t]{0.24\textwidth}  
    \vspace{0pt}
    \centering
    \captionof{table}{\textbf{Quantitative results of FFP via different context sizes.}}
    \vspace{-0.3cm}
    \resizebox{\linewidth}{!}{
    \begin{tabular}{lccc}
      \toprule
      Count & MSE & SSIM & PSNR \\
      \midrule
      1 & 0.287 & 0.449 & 5.63 \\
      2 & 0.262 & 0.464 & 6.13 \\
      4 & 0.249 & 0.568 & 6.36 \\
      8 & 0.236 & 0.627 & 6.59 \\
      16 & 0.217 & 0.651 & 6.90 \\
      \bottomrule
    \end{tabular}
    }
    \label{tab:ablation1}
  \end{minipage}
\end{figure*}

\begin{figure*}[t]
  \centering
  \begin{minipage}[t]{0.49\textwidth}  
    \vspace{0pt}
    \centering
    \includegraphics[width=\textwidth]{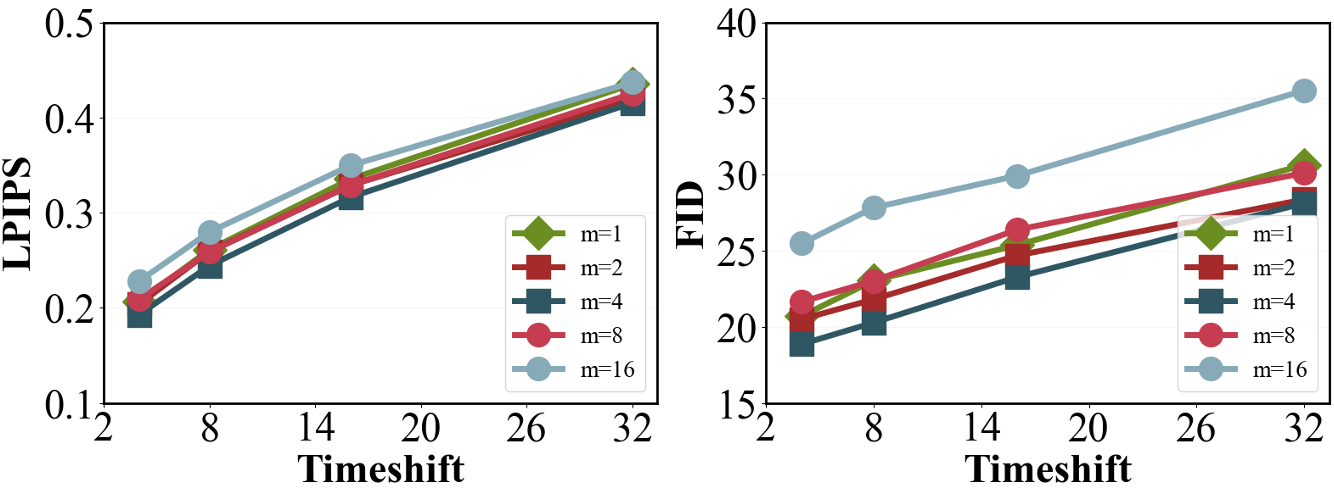}
    \vspace{-0.7cm}
    \caption{\textbf{Ablation of context size for generation}.}
    \label{fig:context_size}
  \end{minipage}
  \hfill
  \begin{minipage}[t]{0.49\textwidth}  
    \vspace{0pt}
    \centering
    \includegraphics[width=\textwidth]{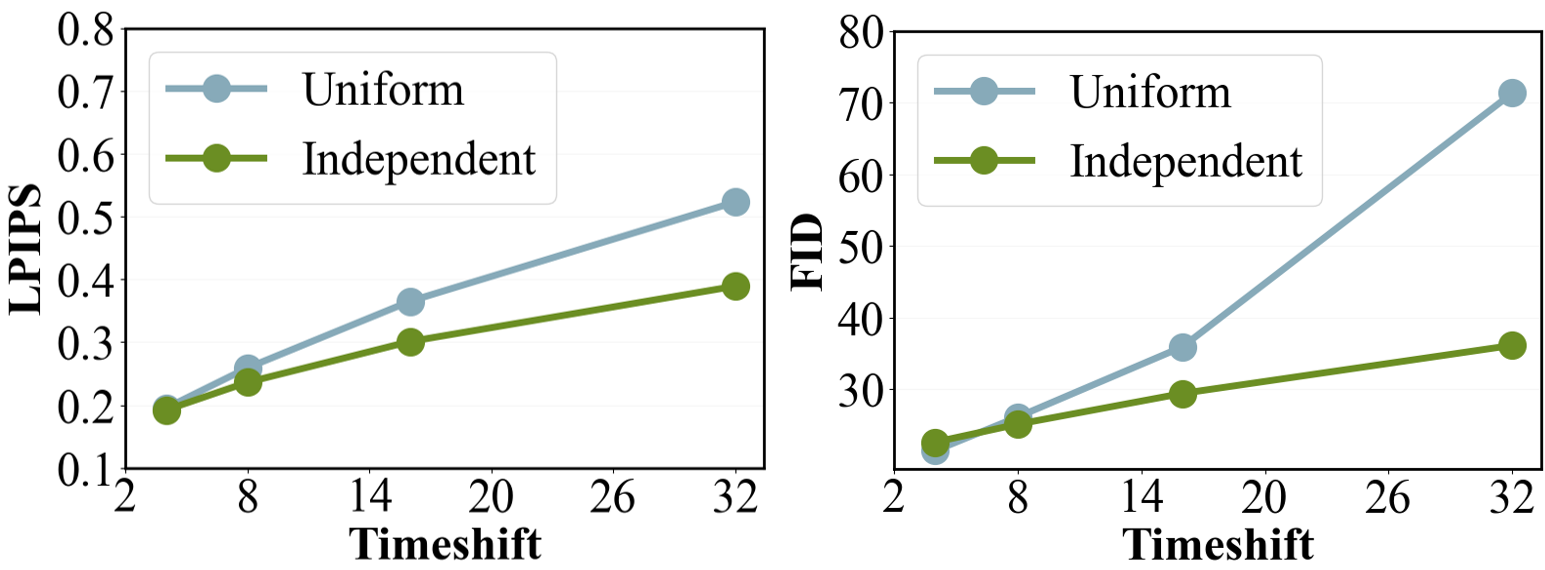}
    \vspace{-0.7cm}
    \caption{\textbf{Ablation of condition latents modulation.}}
    \label{fig:modulation}
  \end{minipage}
  \vspace{-0.4cm}
\end{figure*}

\subsection{Main Results}
\paragraph{Visual Generation}
We report the generation results at 4s, 8s, 16s, and 32s in Table~\ref{tab:gen_2d3d}. 
We have the following observations:
1) For both 2D and 3D trajectories, the performance of all baselines degrades as the trajectory length increases, indicating that the reliability of generated observations decreases with distance.
2) Within the 32s, our method consistently achieves the best performance of generating visual observations across all evaluation metrics for both 2D and 3D trajectories.
3) The results of YUME and NWM at 16 s and 32 s are significantly worse than those of our method, suggesting that they are limited in generating consistent visual observations along long-range trajectories.
Through comparison with these baselines, we conclude that 1) ANWM better captures the correspondence between actions and visual observations; 2) propagating historical scene information during future-frame generation effectively improves the long-horizon generation accuracy of ANWM.

We also present the qualitative results in Figure~\ref{fig:qual_res}. The observations generated by our method are more consistent with the ground truth and exhibit higher visual realism. Although NWM and Matrix-Game can produce visually plausible images, their results gradually deviate from the actual motion trajectory as the path length increases.
In contrast, YUME suffers from mode collapse at the early stage of generation. Even for 3D trajectories with large altitude variations, ANWM can maintain accurate correspondence between the generated observations and the underlying motion trajectory.

\noindent\textbf{Navigation} As depicted in Table~\ref{tab:main_nav}, ANWM achieves the highest navigation success rate and the lowest navigation error in both 2D and 3D navigation tasks. Specifically, the ATE of ANWM is reduced by 5.1\% compared to the second-best method, while the SR is improved by 10\%
For 3D navigation results, our method outperforms NWM by 2\% in terms of SR and 4.7\% in terms of ATE, further demonstrating the effectiveness and robustness of our approach in long-range navigation.

The heuristic path planner first generates 5 candidate trajectories, and ANWM ranks each trajectory’s final prediction by measuring LPIPS similarity with the goal image. The ranking results is demonstrated in Figure~\ref{fig:qual_nav}.

\subsection{Ablation Study}
\noindent\textbf{Context Size for Future Frame Projection} In this section, we project varying number of past frames from 1 to 16 for Future Frame Projection. As illustrated in Figure~\ref{fig:ablation1} and Table~\ref{tab:ablation1}, increasing the number of past frames consistently improves the generation quality of both short-term (4s) and long-term (32s). The visualized results in Figure~\ref{fig:ablation1_qual} and their metrics in Table~\ref{tab:ablation1} demonstrate that incorporating more past frames provides richer contextual information about the environment, enabling the projected future frames to more closely approximate the ground truth. Consequently, the world model benefits from more accurate prior information, leading to higher-quality generation outcomes.

\noindent\textbf{Context Size for Generation} We train the model with different context sizes $m$. To exclude the influence of the projected future frame, we only use the current frame $\mathbf{v}_t$ to generate future frame projection. The results are shown in Figure~\ref{fig:context_size}.
Although prior work~\cite{bar2025navigation} suggests that increasing the context size improves generative performance when the context size is not bigger than 4, our experiments show that the performance degrades when the context size is extended to 16. We assume this is because distant historical frames differ significantly from the future frame, introducing additional noise to the generation process.

\noindent\textbf{Modulation Method for Condition Latents} We train the model with both uniform and independent latent modulation architectures. The former modulates the condition latents of past frames and the projected future frame using the same scale and shift parameters, while the latter uses two separate modulation modules.
The results shown in Figure~\ref{fig:modulation} indicate that the two methods perform comparably in short-range generation. However, for long-horizon generation, the independent modulation significantly outperforms the uniform modulation.

\section{Limitations}
\begin{figure}[t]
  \vspace{-0.3cm}
  \centering
  \includegraphics[width=\linewidth]{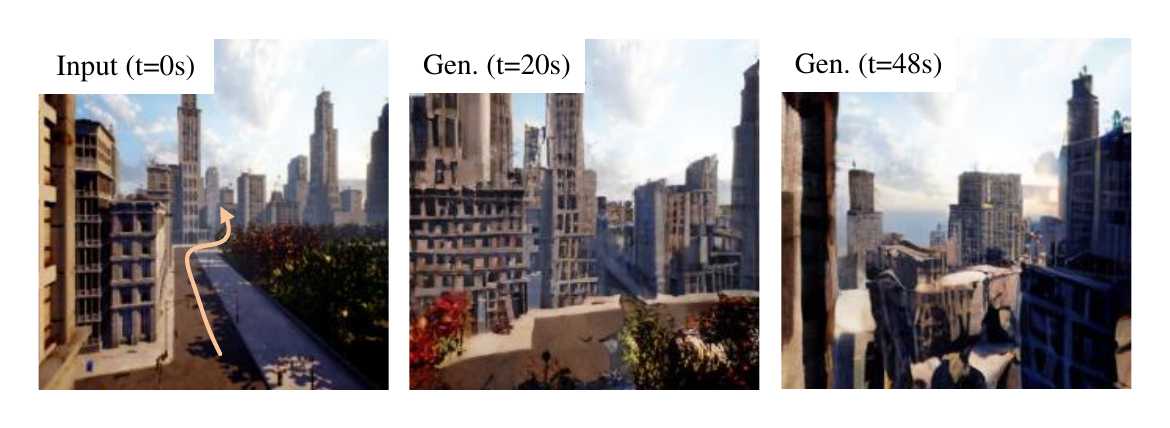}
  \vspace{-0.8cm}
  \caption{\textbf{Limitations of our model.} ANWM fails to generate fine-grained architectural textures (middle) and consistent observations for extremely long-range trajectories (right).}
  \vspace{-0.5cm}
  \label{fig:limitation}
\end{figure}

In this section, we identify several limitations of our approach. While our method can generate realistic visual observations along trajectories of approximately 100 meters, it tends to experience mode collapse~\cite{thanh2020catastrophic,srivastava2017veegan} when extended to longer distances (e.g., around 200 meters). We hypothesize that the accumulated viewpoint variations over such long trajectories lead to large discrepancies between future and historical observations, making the past frames ineffective as priors for future frame prediction.
Second, our method occasionally produces distortions in the generation of fine-grained texture, such as in the details of the windows or facades of the buildings in Figure~\ref{fig:limitation}. To alleviate this issue, we plan to incorporate additional physical constraints to enhance the model’s perception of three-dimensional spatial structure.
Besides, in our current navigation experiments, the world model is primarily used to rank different trajectory candidates. In future work, we plan to enable the world model to assist the UAV in actively planning its own paths.

\section{Conclusion}

In this work, we present ANWM, the first aerial world model capable of generating long-horizon visual observations along 3D trajectories for UAV navigation. Experimental results demonstrate the effectiveness of ANWM in long-range visual generation and 3D navigation accuracy. 
We also discuss several limitations of ANWM in generating fine-grained textures and providing timely guidance for UAV path planning, which we plan to address through enhanced physical constraints and 3D path planning algorithms. 


{
    \small
    \bibliographystyle{ieeenat_fullname}
    \bibliography{main}
}



\end{document}